%% file: main.tex
\documentclass[12pt,oneside,a4paper]{article}

\usepackage[T1]{fontenc}
\usepackage[utf8]{inputenc}
\usepackage[english]{babel}
\usepackage{lmodern}
\usepackage{microtype}
\usepackage{amsmath, amssymb, mathtools}
\usepackage{setspace}
\usepackage{geometry}
\usepackage{float}
\usepackage{tabularx}
\usepackage{graphicx}
\usepackage{booktabs}
\usepackage{xcolor}
\usepackage[hidelinks]{hyperref} 
\usepackage[square,numbers]{natbib}

\title{\textbf{Modular Linear Tokenization (MLT)}}

\author{
    \textbf{Tcharlies Schmitz}\\[4pt]
    \small Data Science — PX.Center\\
    \small \texttt{tcharlies.schmitz@px.center}\\[4pt]
    \small ORCID: \href{https://orcid.org/0009-0007-5467-1327}{0009-0007-5467-1327}\\[4pt]
    \small DOI: \href{https://doi.org/10.5281/zenodo.17467914}{10.5281/zenodo.17467914}
}

\date{October 28, 2025}

\begin{document}
\selectlanguage{english}

\maketitle


\input{abstract}
\input{introduction}
\input{methodology}
\input{results}
\input{discussion}

\input{conclusion}
\input{acknowledgments}

\bibliography{references}

\end{document}

%% file: abstract.tex

\begin{abstract}
    \noindent
    \textbf{Abstract.} 
    This paper introduces \textit{Modular Linear Tokenization} (MLT), a reversible and deterministic technique for encoding high-cardinality categorical identifiers into compact numerical vectors. 
    Unlike traditional hashing or one-hot encodings, MLT preserves bijective mappings by leveraging modular arithmetic over finite fields and invertible linear transformations. 
    The method offers explicit control of dimensionality and computational scalability while maintaining full reversibility, even for millions of identifiers. 
    Experimental results on the \textit{MovieLens 20M} dataset show that MLT achieves comparable predictive performance to supervised embeddings while requiring significantly fewer parameters and lower training cost. 
    An open-source implementation of MLT is available on PyPI (\url{https://pypi.org/project/light-mlt/}) and GitHub (\url{https://github.com/tcharliesschmitz/light-mlt}).
    
    \vspace{0.5em}
    \noindent
    \textbf{Keywords:} Modular Linear Tokenization, reversible encoding, categorical data, finite fields, efficient preprocessing.
    \end{abstract}
    

%% file: introduction.tex
\section{Introduction}

The way we represent categorical and textual data is decisive for the success of machine learning models—especially in deep architectures. In high-cardinality scenarios, such as recommendation systems, user/item classification, or large vocabularies, the tokenization stage often becomes a bottleneck: it consumes large amounts of memory, increases computational cost, and limits generalization capacity.

This work presents the \textit{Modular Linear Tokenization} (MLT), a deterministic technique that transforms integer identifiers into compact vectors while preserving exact reversibility. The proposal combines three key advantages: (i) explicit control over dimensionality, (ii) a significant reduction in output cost for tasks with a large number of labels, and (iii) the absence of collisions—something not guaranteed by methods such as \textit{hashing}.

Historically, the first approaches date back to the 1950s–1960s with one-hot and bag-of-words (BoW) representations \cite{harris1954distributional}. These encodings are simple but produce extremely large and sparse vectors whose size grows linearly with the vocabulary.

In the 1970s, TF–IDF introduced a statistical notion to frequency counts but maintained high dimensionality \cite{sparckjones1972statistical}.
Between the 1980s and 1990s, n-grams and count-based models emerged, capturing local context but suffering from combinatorial explosion for large n \cite{chen1996empirical}.

In the 2000s, within the tabular domain, encodings such as \textit{label encoding} (fast but prone to artificial ordering) and supervised variants such as \textit{target}/\textit{count} encoding (more informative but subject to leakage) appeared. 
In 2009, the hashing trick offered a way to project categories into a fixed-dimensional space, stabilizing production pipelines but introducing collisions and losing reversibility \cite{weinberger2009feature}.

The following decade consolidated learned distributed embeddings, such as word2vec (2013) \cite{mikolov2013efficient} and GloVe (2014) \cite{pennington2014glove}, which captured semantics in dense vectors. fastText (2016–2017) incorporated morphology through character n-grams \cite{joulin2016bag}.

In NLP, tokenization evolved from whole words to subwords: BPE (2015) \cite{sennrich2015neural} and WordPiece (2016) \cite{wu2016google} balanced robustness and coverage, followed by SentencePiece (2018), which unified vocabularies and normalization \cite{kudo2018sentencepiece}. 
From 2019 onward, byte-level variants (e.g., GPT-2) and tokenizer-free models (ByT5, 2021) sought to simplify vocabularies \cite{xue2021byt5}. 

Between 2022 and 2024, engineering practices consolidated unified normalizers, compact multilingual vocabularies, and high-performance tokenizers, while recent research explores hybrid schemes, domain specialization, and even reversible representations.\footnote{See recent proceedings from ACL, EMNLP, NAACL conferences (2022-2024) for in-depth studies on state-of-the-art tokenizers and normalizers.}

Despite these advances, three gaps remain:
\begin{enumerate}
    \item the lack of reversibility and the risk of collisions in efficient methods such as \textit{hashing};
    \item the absence of explicit control over the relationship between dimensionality and cost;
    \item the difficulty of reducing the cost of output layers when the number of classes is very large.
\end{enumerate}

MLT positions itself precisely in this space. By representing identifiers in base-$p$ (with $p$ prime) and applying invertible modular linear transformations, it guarantees a bijective mapping, direct control of dimensionality, and a natural decomposition of the output into $n$ \textit{softmax} layers of size $p$, instead of a single \textit{softmax} of size $V$.

The objective of this paper is to formalize MLT, present its properties—uniqueness, reversibility, and dimensionality control—and discuss its practical impact on training and inference costs in high-cardinality scenarios. The contributions include: the mathematical formulation and encoding/decoding algorithms, guidelines for selecting $p$ and $n$ according to the vocabulary size, and the analysis of its integration into deep architectures, with emphasis on output cost reduction.

%% file: methodology.tex
\section{Methodology}

The idea behind \textit{Modular Linear Tokenization} (MLT) is straightforward: to transform high-cardinality integer identifiers (such as user or contract IDs) into smaller vectors without losing the ability to recover the original value. Unlike \textit{hashing} techniques, which may produce collisions, MLT is fully reversible. To achieve this, the method relies on three pillars: (i) representation in base-$p$, (ii) the application of an invertible matrix over this vector, and (iii) the guarantee of uniqueness in representations.

The starting point is the conversion of an identifier into base-$p$. Instead of handling a large integer, the value is decomposed into a fixed-length vector of digits. Each digit takes values between $0$ and $p-1$, with $p$ chosen as a prime number to facilitate subsequent operations. For this representation to be sufficient, the space of combinations must be larger than the vocabulary size $V$ (i.e., the maximum number of categories to represent):

\begin{equation}
  p^{n} > V
\end{equation}

Next comes the matrix $M$, defined over the finite field $\mathbb{Z}_{p}$. This matrix must be invertible, meaning that its determinant is nonzero modulo $p$:

\begin{equation}
  \det(M) \not\equiv 0 \pmod{p}
\end{equation}

The role of the matrix is to densely mix the digits, spreading information across the vector’s positions. This prevents numerically close identifiers from producing similar representations, thus enriching the encoding. In practice, each final component of the vector depends on multiple original positions, ensuring greater robustness.

The tokenization process therefore follows a simple sequence:
\begin{enumerate}
  \item convert the identifier to base-$p$;
  \item multiply the resulting vector by matrix $M$;
  \item reduce the result modulo $p$.
\end{enumerate}

The final token vector is defined as:
\begin{equation}
  t = (M \cdot v) \bmod p
\end{equation}

This vector can then be used directly as input to machine learning models.

Decoding consists of applying the inverse process. The inverse matrix $M^{-1}$ is computed in $\mathbb{Z}_{p}$, and the inverse transformation is applied:

\begin{equation}
  v = (M^{-1} \cdot t) \bmod p
\end{equation}

Finally, the vector $v$ is reconverted from base-$p$ into its integer form, recovering the original identifier. In practical terms, encoding can be summarized as: convert $\rightarrow$ multiply $\rightarrow$ reduce modulo $p$ $\rightarrow$ obtain tokens. Decoding follows the reverse path. Both steps involve only basic modular arithmetic operations, making them fast and scalable even in extremely high-cardinality scenarios.

Among the most relevant properties of MLT are:
\begin{itemize}
  \item \textbf{Reversibility}: ensures a bijective mapping between IDs and token vectors, with no collisions;
  \item \textbf{Dimensional control}: parameters $p$ and $n$ allow explicit calibration of computational cost as a function of vocabulary size;
  \item \textbf{Efficiency}: encoding and decoding involve only simple operations, enabling large-scale deployment of the technique.
\end{itemize}

Regarding computational complexity, \textit{Modular Linear Tokenization} (MLT) is linear with respect to the number of digits $n$ in the base-$p$ representation, that is, $\mathcal{O}(n)$. Each encoding or decoding operation involves only the conversion of the identifier into a fixed-length vector and a vector–matrix multiplication modulo $p$. Since both steps depend solely on $n$ and not on the total vocabulary size $V$, the cost remains proportional to the vector’s dimension, regardless of the overall cardinality of the categorical space.

%% file: results.tex

\section{Results}
\label{sec:results}

This section presents the experimental results obtained using the \textit{MovieLens 20M} dataset, one of the most widely used large-scale benchmarks in recommender system research, comprising approximately 20 million user--movie interactions. The main objective was to evaluate the effectiveness of the \textit{Modular Linear Tokenization} (MLT) technique compared to traditional high-cardinality encoding methods such as \textit{One-Hot Encoding} and the \textit{Hashing Trick}, as well as the \textit{MLT+Autoencoder} variant, in the task of predicting whether users assign ratings higher than 4 stars to movies---with both users and items represented solely by their \textit{IDs}. Additionally, we included as an independent \textit{baseline} a model based on supervised \textit{embeddings}, in order to provide a consolidated reference for expected performance in contemporary binary rating classification tasks.

The evaluation considered multiple aspects of the representations: the final vector dimension, reversibility, number of learned parameters, average training time per epoch, inference time per sample, and the accuracy achieved on the binary prediction task. These indicators allow for the observation of not only predictive quality but also efficiency and scalability across different approaches.

Table~\ref{tab:comparativo} summarizes the quantitative results of the fixed encoding methods. The \textit{One-hot} approach achieved competitive performance (74.18\% accuracy) but at the cost of extremely large vectors (164,320 dimensions) and a very high training time. The \textit{Hashing Trick}, in contrast, reduced dimensionality to 512 positions and training time, but with a substantial drop in accuracy (57.87\%). The pure MLT method achieved 63.31\% accuracy using only 14 dimensions and no learned parameters, demonstrating strong computational efficiency, while the \textit{MLT+Autoencoder} variant recovered part of the lost performance, reaching 62.53\% accuracy with vectors of only 16 dimensions.

Table~\ref{tab:embeddings} presents the supervised \textit{Embeddings} baseline, which achieved 74.4\% accuracy---comparable to \textit{One-hot} performance, but at the cost of over 5 million trainable parameters. This illustrates the trade-off between predictive performance and computational efficiency, positioning MLT as a practical alternative in scenarios constrained by memory and latency requirements.

All experiments were implemented using the open-source repository available at \citep{schmitz2025repo}, which provides full training scripts, evaluation routines, and preprocessed datasets to ensure reproducibility. The core implementation of the \textit{Modular Linear Tokenization} is distributed as a Python package under the name \texttt{light-mlt} \citep{schmitz2025pypi}, accessible via the PyPI registry.

\begin{table}[t]
    \centering
    \scriptsize
    \setlength{\tabcolsep}{4pt}
    \renewcommand{\arraystretch}{1.1}
    \resizebox{\linewidth}{!}{
    \begin{tabular}{@{}lcccccc@{}}
    \toprule
    \textbf{Method} & \textbf{Dimension} & \textbf{Reversible} & \textbf{Parameters} & \textbf{Training (s/epoch)} & \textbf{Inference (\(\mu\)s)} & \textbf{Accuracy (\%)} \\
    \midrule
    One-hot        & 164{,}320 & Yes & 0      & 5029.90 & 427.03 & 74.18 \\
    MLT            & 14        & Yes & 0      & 65.04   & 19.93  & 63.31 \\
    MLT+Autoenc.   & 16        & Yes (MLT+Decoder) & 79{,}758 & 2.13 & 14.35 & 62.53 \\
    Hashing        & 512       & No  & 0      & 144.27  & 86.74  & 57.87 \\
    \bottomrule
    \end{tabular}}
    \caption{Comparative results between different fixed encoding methods on the MovieLens 20M dataset.}
    \label{tab:comparativo}
\end{table}

\begin{table}[t]
    \centering
    \scriptsize
    \setlength{\tabcolsep}{4pt}
    \renewcommand{\arraystretch}{1.1}
    \resizebox{\linewidth}{!}{
    \begin{tabular}{@{}lcccccc@{}}
    \toprule
    \textbf{Method} & \textbf{Dimension} & \textbf{Reversible} & \textbf{Parameters} & \textbf{Training (s/epoch)} & \textbf{Inference (\(\mu\)s)} & \textbf{Accuracy (\%)} \\
    \midrule
    Supervised Embeddings & 32 & No & 5{,}287{,}648 & 410.55 & 22.56 & 74.40 \\
    \bottomrule
    \end{tabular}}
    \caption{Results for the supervised \textit{Embeddings} baseline on the MovieLens 20M dataset.}
    \label{tab:embeddings}
\end{table}

%% file: discussion.tex
\section{Discussion}

The results demonstrate that \textit{Modular Linear Tokenization} (MLT) effectively balances interpretability, scalability, and reversibility in categorical data representation. Its linear computational complexity ensures that the encoding process remains predictable even in large-scale scenarios involving millions of identifiers—a distinct advantage over methods such as one-hot encoding, whose cost increases proportionally to the vocabulary size $V$. This characteristic makes MLT particularly suitable for production environments where latency and memory constraints are critical.

Although the accuracy achieved by MLT does not yet surpass that of supervised embeddings, its deterministic and collision-free structure provides benefits that go beyond predictive performance. Unlike embeddings, which rely on stochastic training and may vary across runs, MLT produces stable, reproducible encodings—an essential property for traceable machine learning pipelines and explainable AI applications.

The integration of lightweight compression mechanisms, such as autoencoders, further enhances MLT’s potential by closing part of the performance gap while preserving its reversibility. This hybrid approach suggests a promising research direction for combining deterministic encoding with representation learning, enabling models that are both efficient and theoretically grounded.

In summary, MLT emerges as a practical alternative for representing high-cardinality categorical variables when reproducibility, compactness, and reversibility are as relevant as accuracy. Future work should explore adaptive parameterization of $p$ and $n$, as well as applications of MLT in domains beyond recommender systems—particularly in tabular and graph-based learning settings.

%% file: conclusion.tex
\section{Conclusion}

The Modular Linear Tokenization (MLT) framework introduces a novel paradigm for representing categorical data and integer identifiers in machine learning architectures. Grounded in modular arithmetic and invertible linear transformations over finite fields, MLT provides a fully reversible and collision-free encoding with explicit control over dimensionality—three properties seldom achieved simultaneously by conventional methods. This mathematical structure grants MLT a unique robustness, positioning it as an intermediary approach between deterministic encodings and learned embeddings.

Experimental evidence demonstrates that MLT significantly reduces computational cost while preserving competitive predictive performance, even without learned parameters. Its algorithmic simplicity and predictable linear complexity make it particularly suitable for large-scale tabular models, recommendation systems, and high-cardinality classification pipelines, where efficiency and auditability are paramount.

Beyond its empirical utility, MLT offers a theoretical foundation that invites further exploration. Its modular and invertible nature enables hybrid architectures that combine deterministic tokenization with learned compression, integration into neural embedding layers, and potential applications in vector indexing and deterministic embedding compression. This opens a promising research direction in which categorical representation is treated not as a preprocessing step but as a mathematically principled component of model design.

In summary, MLT demonstrates that \textbf{efficiency, reversibility, and interpretability} can coexist within a unified mathematical formulation. By bridging symbolic encoding and continuous representation learning, it contributes to the development of more transparent, efficient, and theoretically grounded machine learning systems. 

Future work includes exploring hybrid models that combine MLT with learned embeddings and evaluating its impact on large-scale industrial recommendation pipelines.

%% file: acknowledgments.tex
\section*{Acknowledgments}
This research was supported by \textbf{PX.Center} --- a Brazilian logistics platform focused on freight brokerage and transportation optimization (\url{https://px.center}). 
The PX.Center provided computational infrastructure, datasets, and a research environment that enabled the development and validation of the \textit{Modular Linear Tokenization (MLT)} methodology.